%% file: main.tex
\documentclass[letterpaper]{article} 
\usepackage[draft]{aaai2027}  
\usepackage[hyphens]{url}  
\usepackage{graphicx} 
\usepackage{natbib}  
\usepackage{caption} 
\usepackage{algorithm}
\usepackage{algorithmic}
\usepackage{enumitem}

\usepackage{booktabs}
\usepackage{multirow}
\usepackage{graphicx}
\usepackage[table]{xcolor}
\usepackage{amsmath}
\usepackage{amssymb}
\usepackage{amsfonts}
\usepackage{bm}

\definecolor{sectiongray}{RGB}{242,242,242}
\definecolor{oursblue}{RGB}{232,244,248}
\definecolor{selectedred}{RGB}{149,55,52}

\definecolor{codeblue}{RGB}{48,88,145}
\definecolor{codeurl}{RGB}{45,82,135}

\newcommand{\best}[1]{\textbf{#1}}
\newcommand{\second}[1]{\underline{#1}}

\newcommand{\selectedcoef}[1]{#1\textsuperscript{\(\dagger\)}}

\newcommand{\grouphead}[2]{%
    \multicolumn{13}{c}{%
        \rule{0pt}{2.5ex}\textbf{\textit{(#1) #2}}%
    }\\[-0.2ex]
    \cmidrule(lr){1-13}
}

\newcommand{\mymethod}{STEP-OPD}

\usepackage{newfloat}
\usepackage{listings}
\DeclareCaptionStyle{ruled}{labelfont=normalfont,labelsep=colon,strut=off} 
\floatstyle{ruled}
\newfloat{listing}{tb}{lst}{}
\floatname{listing}{Listing}

\usepackage{booktabs}

\title{STEP-OPD: Rethinking Output Targets and Internal Dynamics in On-Policy Distillation for Diffusion Models}
\author{
    Qingyan Wei\textsuperscript{\rm 1}\equalcontrib,
    Guangzhao Li\textsuperscript{\rm 1, \rm 3}\equalcontrib, Xiaobing Tu\textsuperscript{\rm 2}, Yinggui Wang\textsuperscript{\rm 2}, Xiantao Zhang\textsuperscript{\rm 2}, Jinkui Ren\textsuperscript{\rm 2}, Xiaohong Liu\textsuperscript{\rm 1, \rm 3}, Linfeng Zhang\textsuperscript{\rm 1}\corresponding
}
\affiliations{
    \textsuperscript{\rm 1}Shanghai Jiao Tong University
    \quad
    \textsuperscript{\rm 2}Alibaba Group 
    \quad
    \textsuperscript{\rm 3}Shanghai Innovation Institute
    \\[4pt]

    \textcolor{codeblue}{\textbf{Homepage:}}
    \url{https://stepopd.github.io}
}

\begin{document}

\maketitle

\input{section/abstract}
\input{section/intro}
\input{section/related_work}
\input{section/method}

\input{section/experiments}
\input{section/conclusion}


\newpage
\bibliography{aaai2027}


\end{document}

%% file: section/abstract.tex
\begin{abstract}
On-policy distillation (OPD) has become an effective approach for consolidating multiple task-specialized image generation models into a single student. However, existing OPD methods optimize the student mainly to match the teacher's output velocity, making the teacher the upper limit of the optimization objective. While output-level supervision alone leaves the student's blockwise representation evolution underconstrained, which weakens the transfer of capabilities that must be progressively developed across layers. We propose \textbf{\mymethod}, an on-policy distillation framework for image generation that extends the student's learning target beyond the teacher and introduces explicit constraints on its internal representation evolution. Instead of treating the teacher as the final target, we use the velocity difference between each task-specific teacher and the shared base model as a direction for further learning and add a scaled version of this difference to the teacher velocity. In addition, we align the direction and magnitude of representation changes between the student and teacher, enabling the student to learn how representations are progressively transformed across network blocks. Experiments on compositional alignment, text rendering, and human preference show that our method consistently improves Standard OPD methods. In particular, it increases the GenEval score of DiffusionOPD from $\textbf{0.927}$ to $\textbf{0.961}$, while also improving OCR and all preference-based metrics. The resulting unified student surpasses the corresponding single-task teachers across all three capability groups, showing that output extrapolation enables beyond-teacher learning. And representation change alignment provides complementary guidance for the student’s internal transformations.
\end{abstract}

%% file: section/intro.tex
\section{Introduction}

\begin{figure}[t]
    \centering
    \includegraphics[width=1.0\linewidth]{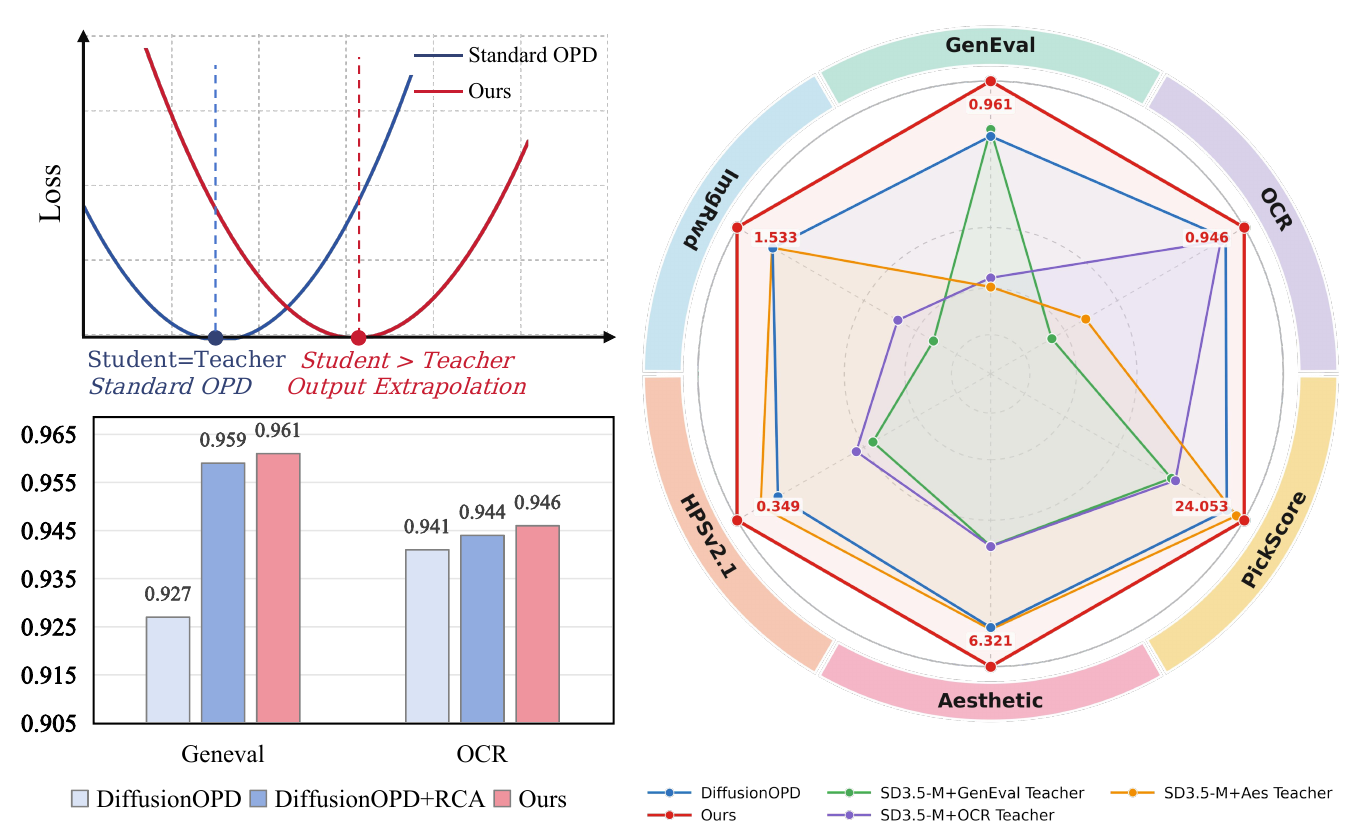}
    \caption{\textbf{Motivation and overview of our method.} \textbf{Upper Left:} Standard OPD reaches its optimum at exact teacher matching, whereas output extrapolation shifts the learning target beyond the teacher. \textbf{Lower Left:} Representation change alignment improves GenEval and OCR without changing the output supervision target, supporting the need for internal representation supervision. \textbf{Right:} Our method achieves the strongest overall performance across compositional alignment, text rendering, and human preference.}
    \label{fig:teaser}
\end{figure}

On-policy distillation (OPD)~\cite{agarwal2024policy, song2026survey,coreteam2026mimov2flashtechnicalreport,zhou2026danceopd,li2026diffusionopd, fang2026flow} has recently emerged as an effective post-training approach for image generation. Recent studies show that the capabilities of multiple task-specialized diffusion or flow models can be consolidated into a single student model through supervision on the student’s own denoising trajectories. This paradigm separates expert training based on a single reward from multi task capability integration, thereby resolving reward conflicts and optimization imbalances~\cite{chen2026mapreduceloraadvancingpareto}. Moreover, by querying teachers at states visited by the student, OPD for image generation provides dense supervision throughout the denoising process and reduces the mismatch between training and inference trajectories.

However, existing OPD for image generation still has two important limitations. First, it is limited to the objective of teacher matching. The student is optimized only to match the teacher’s velocity at each visited state, making the teacher itself the upper bound of learning. As illustrated in the upper-left panel of Fig.~\ref{fig:teaser}, for a fixed denoising state, the standard squared velocity-matching loss can be viewed as a quadratic function of the student's relative position along the teacher-induced direction. Its ideal minimum is attained exactly when the student prediction matches the teacher prediction. Moving beyond the teacher instead increases the loss and pushes the student back toward teacher matching. Therefore, standard OPD does not provide an optimization target for further improvement beyond it. Even with perfect optimization, the student cannot surpass the teacher under this objective. 


Second, output velocity matching provides only an endpoint signal. Since the final velocity aggregates the effects of all network blocks, deviations in the representation updates of different layers may offset each other in the final prediction~\cite{yu2025representationalignmentgenerationtraining, yim2017gift}. The student can therefore achieve a low output loss through this compensation without faithfully learning the teacher's structured processing of visual tokens and conditioning information. This weakens the transfer of capabilities that must be progressively developed across layers, such as attribute binding, spatial composition, and text rendering. As shown in the lower-left panel of Fig.~\ref{fig:teaser}, adding representation change alignment to DiffusionOPD, substantially improves GenEval and OCR even when the output target remains unchanged. This controlled comparison supports our analysis that output matching alone leaves the student's internal representation evolution underconstrained, and that directly supervising blockwise representation changes enables more effective transfer of the teacher's compositional and text-rendering capabilities.

To address these limitations, we propose \textbf{\mymethod}, an on-policy distillation framework for image generation that extends the student's learning target beyond the teacher and provides direct supervision for how representations evolve inside the network. For the first limitation, we introduce output extrapolation. Instead of treating the teacher as the final learning target, our method uses the velocity difference between each task teacher and the shared base model to define a direction for further learning, and constructs a target beyond the teacher along this direction by adding it to the teacher velocity. For the second limitation, we introduce representation change alignment, which matches the direction and magnitude of hidden-state changes between consecutive blocks in the student and teacher. 


We conduct extensive experiments across compositional alignment, text rendering, and human preference. As shown in the right panel of Fig.~\ref{fig:teaser}, our method achieves the strongest overall capability profile compared with DiffusionOPD and the task-specialized teachers, with consistent advantages across compositional alignment, text rendering, and human preference.  In particular, it increases the GenEval overall score of DiffusionOPD from 0.927 to 0.961, while also improving all human-preference metrics. More importantly, it surpasses the corresponding single-task teacher in each capability category, demonstrating effective capability consolidation and beyond-teacher performance.

Our contributions are summarized as follows:

\begin{itemize}
    \item We propose \textbf{\mymethod}, an on-policy distillation framework for image generation that enables the student to learn beyond teacher matching while transferring the teacher's internal representation dynamics.

    \item We introduce output extrapolation, which adds a scaled difference between each task teacher and the shared base model to the teacher velocity, producing a learning target beyond the teacher. We further propose intermediate representation change alignment, which matches the direction and magnitude of blockwise representation updates between the student and teacher.

    \item Extensive experiments show that our method significantly outperforms both DanceOPD and DiffusionOPD. In particular, it increases the GenEval score of DiffusionOPD from 0.927 to 0.961 and enables the unified student to surpass the corresponding single-task teachers across all three capability categories.
\end{itemize}

%% file: section/related_work.tex
\section{Related Work}

\noindent\textbf{On-Policy Distillation for Image Generation.}
On-policy distillation trains the student on its own denoising trajectories and queries the teacher at the states visited by the current student. Recent works have extended this paradigm to image generation: DiffusionOPD formulates OPD for diffusion processes, Flow-OPD applies it to multi-task alignment of flow-matching models, DanceOPD distills multiple generative fields through task routing, and D-OPSD studies on-policy self-distillation for continually tuning few-step diffusion models \cite{li2026diffusionopd,fang2026flow,zhou2026danceopd,jiang2026d}. These methods reduce the mismatch between training and inference trajectories and provide dense supervision throughout generation. However, they mainly optimize the student to reproduce the teacher’s output field, whereas our method further constructs targets beyond the teacher and introduces representation-level supervision.

\medskip

\noindent\textbf{Guidance in Diffusion and Flow Models.}
Guidance methods steer diffusion generation by modifying the predicted score or velocity field. Classifier guidance uses gradients from an external classifier, while classifier-free guidance combines conditional and unconditional predictions \cite{dhariwal2021diffusion,ho2022classifier}. Guided Flows extends classifier-free guidance to flow-matching models, Autoguidance constructs a guidance direction from a main model and a weaker version of itself, and subsequent work further studies guidance for flow matching \cite{zheng2023guided, karras2024guiding, feng2025guidance}. These studies show that differences between prediction fields can provide useful directions for generation. Unlike inference-time guidance, our method uses the post-RL teacher–base velocity difference during training to construct improved on-policy distillation targets for a single student.

\medskip

\noindent\textbf{Representation-Level Distillation.}
Knowledge distillation was initially developed at the output level and was later extended to intermediate features, attention maps, and relations between representations \cite{hinton2015distilling,romero2014fitnets,zagoruyko2016paying,park2019relational}. For diffusion models, BK-SDM uses feature distillation to train compact Stable Diffusion models, while KOALA emphasizes self-attention features for efficient text-to-image generation \cite{kim2024bk, lee2024koala}. OPRD further introduces hidden-state alignment on on-policy trajectories for language models \cite{yang2026oprd}. Unlike prior methods that directly match intermediate representations, our method aligns how representations evolve across the network by matching the direction and magnitude of inter-layer hidden-state changes between the student and teacher.

%% file: section/method.tex
\section{Method}

\begin{figure*}[t]
    \centering
    \includegraphics[width=1.0\linewidth]{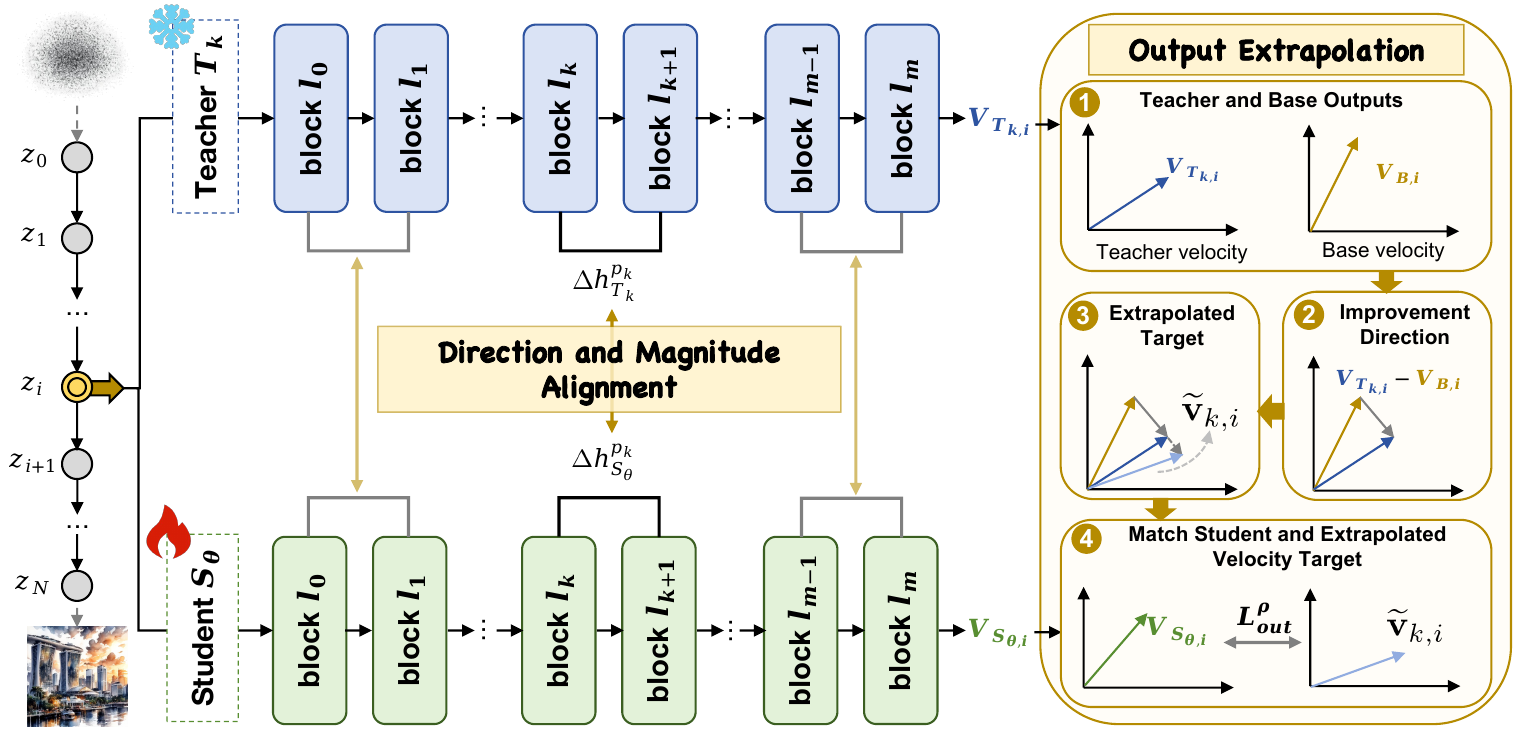}
\caption{\textbf{Overview of STEP-OPD.}
The student generates an on-policy denoising trajectory, and a latent state $\mathbf{z}_i$ is selected according to the underlying OPD query strategy.
\textbf{Left: Representation Change Alignment.}
The routed teacher and student process the same state, while STEP-OPD aligns the direction and magnitude of their hidden-state changes across selected adjacent blocks.
\textbf{Right: Task-Specific Output Extrapolation.}
The velocity difference between the teacher and shared base model defines a task-specific improvement direction, which is scaled and added to the base velocity to construct a beyond-teacher target.
Together, the two objectives extend output learning beyond teacher matching while enabling the student to learn how internal representations progressively evolve across Transformer blocks.}
    \label{fig:method_overview}
\end{figure*}

\subsection{Preliminaries}
\label{sec:preliminaries}

We consider a shared base model $B$, a student model $S_{\theta}$, and $K$ task-specific teachers $\{T_k\}_{k=1}^{K}$, where each teacher is obtained by post-training $B$ for a specific task. 

In on-policy distillation, the current student generates a denoising trajectory $\tau=(\mathbf{z}_0,\ldots,\mathbf{z}_N)$, and the routed teacher is evaluated at the same latent states visited by the student. Let $p_{\theta}(\tau\mid c)$ and $p_{T_k}(\tau\mid c)$ denote the trajectory distributions of the student and teacher, respectively. The trajectory-level reverse KL objective can be decomposed into transition-level KL objectives along the student trajectory:
\begin{equation}
\begin{aligned}
\mathcal{L}_{\mathrm{OPD}}^{\mathrm{traj}}
&=
\mathbb{E}_{k,\,c\sim\mathcal{D}_k}
\left[
D_{\mathrm{KL}}
\left(
p_{\theta}(\tau\mid c)
\,\Vert\,
p_{T_k}(\tau\mid c)
\right)
\right] \\
&=
\mathbb{E}_{\substack{
k,\,c\sim\mathcal{D}_k\\
\tau\sim p_{\theta}(\tau\mid c)
}}
\left[
\sum_{i=0}^{N-1}
D_{\mathrm{KL}}
\left(
q_{\theta,i}
\,\Vert\,
q_{T_k,i}
\right)
\right].
\end{aligned}
\label{eq:opd_trajectory}
\end{equation}

For diffusion and flow-based image generation, a one-step transition can be expressed as
$p_M(\mathbf{z}_{i+1}\mid\mathbf{z}_i,c)
=
\mathcal{N}
\left(
\boldsymbol{\mu}_M(\mathbf{z}_i,t_i,c),
\boldsymbol{\Sigma}_i
\right),
\label{eq:transition_distribution}$ 
where  $M\in\{S_{\theta},T_k\}$, $\boldsymbol{\mu}_M$ is the transition mean, and $\boldsymbol{\Sigma}_i$ is determined by the sampling schedule. When the student and teacher use the same schedule, their transition distributions share the same covariance. The transition-level KL therefore reduces to
\begin{equation}
D_{\mathrm{KL}}
\left(
p_{\theta}
\,\Vert\,
p_{T_k}
\right)
=
\frac{1}{2}
\left\|
\boldsymbol{\mu}_{S_{\theta}}
-
\boldsymbol{\mu}_{T_k}
\right\|_{\boldsymbol{\Sigma}_i^{-1}}^{2}.
\label{eq:transition_kl}
\end{equation}

In the deterministic flow setting, the next state is fully determined by the predicted velocity field. For a solver step from $t_i$ to $t_{i+1}$, the transition mean is
\begin{equation}
\boldsymbol{\mu}_M(\mathbf{z}_i,t_i,c)
=
\mathbf{z}_i
+
\Delta t_i\,
\mathbf{v}_M(\mathbf{z}_i,t_i,c),
\label{eq:flow_transition}
\end{equation}
where $\mathbf{v}_M$ is the conditional velocity predicted by model $M$ and
$\Delta t_i=t_{i+1}-t_i$ is the signed solver step size. Since the student and teacher are evaluated at the same state $\mathbf{z}_i$, their transition-mean difference satisfies
\begin{equation}
    \boldsymbol{\mu}_{S_{\theta}}
-
\boldsymbol{\mu}_{T_k}
=
\Delta t_i
\left(
\mathbf{v}_{S_{\theta}}
-
\mathbf{v}_{T_k}
\right).
\label{eq:mean_velocity_relation}
\end{equation}


Consequently, deterministic transition matching like DiffusionOPD ~\cite{li2026diffusionopd} is equivalent to
velocity matching like DanceOPD ~\cite{zhou2026danceopd} up to the known timestep-dependent factor
$(\Delta t_i)^2$. We absorb this factor, together with any global
normalization constant, into a non-negative timestep weight $w_i$. To accommodate different teacher-query strategies, let
$\rho_k(i\mid c,\tau)$ be a distribution over the states of the current
student trajectory, satisfying $\rho_k(i\mid c,\tau)\geq 0,
\sum_{i=0}^{N-1}\rho_k(i\mid c,\tau)=1.$
We additionally define $\bar{\mathbf{z}}_i
=
\operatorname{sg}(\mathbf{z}_i)$
as the stop-gradient student state. A query-general output-level OPD
objective can be written as
\begin{equation}
\begin{aligned}
\mathcal{L}_{\mathrm{OPD}}^{\rho}
&=
\mathbb{E}_{\substack{
k\sim\pi,\;
c\sim\mathcal{D}_k\\
\tau\sim p_{\theta}(\tau\mid c)
}}
\Biggl[
\sum_{i=0}^{N-1}
\rho_k(i\mid c,\tau)\,
w_i
\Bigl\|
\mathbf{v}_{S_{\theta}}
\left(
\bar{\mathbf{z}}_i,t_i,c
\right)
\\[-0.5ex]
&\hspace{9em}
-
\operatorname{sg}
\left[
\mathbf{v}_{T_k}
\left(
\bar{\mathbf{z}}_i,t_i,c
\right)
\right]
\Bigr\|_2^2
\Biggr]
\\
&=
\mathbb{E}_{\substack{
k\sim\pi,\;
c\sim\mathcal{D}_k\\
\tau\sim p_{\theta}(\tau\mid c),\;
I\sim\rho_k(\cdot\mid c,\tau)
}}
\Biggl[
w_I
\Bigl\|
\mathbf{v}_{S_{\theta}}
\left(
\bar{\mathbf{z}}_I,t_I,c
\right)
\\[-0.5ex]
&\hspace{9em}
-
\operatorname{sg}
\left[
\mathbf{v}_{T_k}
\left(
\bar{\mathbf{z}}_I,t_I,c
\right)
\right]
\Bigr\|_2^2
\Biggr].
\end{aligned}
\label{eq:query_general_opd}
\end{equation}
Here, $\operatorname{sg}(\cdot)$ denotes stop-gradient. The rollout is
generated by the current student and refreshed during training, while
gradients are not propagated through the trajectory-generation solver.
The distribution $\rho_k$ determines where the routed teacher is
queried along the student trajectory, whereas $w_i$ determines the
local contribution of each queried timestep.

Under this formulation, DiffusionOPD is instantiated with the uniform query distribution
$\rho_k(i\mid c,\tau)=1/N$ and the transition-induced weight
$w_i=(\Delta t_i)^2/2$, evaluating all $N$ states along each trajectory. In contrast, DanceOPD uses a semantic query distribution
$\rho_k(i\mid c,\tau)=q_{\mathrm{sem}}(i)$ that assigns greater
probability to states in the low-noise region, sets $w_i=1$, and samples a
single index $I\sim q_{\mathrm{sem}}$ from each trajectory.

This objective provides dense teacher supervision on the student's own trajectory and reduces the mismatch between training and inference states. However, standard OPD limits the student’s optimization target to matching the teacher and relies only on output velocity supervision, providing insufficient guidance on the internal transformations that produce the final prediction. Building on the same on-policy trajectory, we address these limitations through output extrapolation and intermediate representation change alignment.

\subsection{Task-Specific Output Extrapolation}

Since each task-specific teacher is obtained from the same base model, the difference between their velocity predictions reflects the change introduced by post-training. As illustrated in the right panel of Fig.~\ref{fig:method_overview}, instead of using the teacher velocity as the final target, we treat this difference between the base model and the teacher as a direction for further learning.

For task $k$, we construct the extrapolated velocity target as
\begin{equation}
\begin{aligned}
\widetilde{\mathbf{v}}_{k,i}(s)
&=
\mathbf{v}_{T_k,i}
+
\alpha_k(s)
\left(
\mathbf{v}_{T_k,i}
-
\mathbf{v}_{B,i}
\right)
\\[-0.3ex]
&=
\mathbf{v}_{B,i}
+
\left(
1+\alpha_k(s)
\right)
\left(
\mathbf{v}_{T_k,i}
-
\mathbf{v}_{B,i}
\right),
\end{aligned}
\label{eq:velocity_extrapolation}
\end{equation}
where all velocity predictions are evaluated at the same
$(\mathbf{z}_t,t,c)$, and $\alpha_k(s) \geq 0$ is a fixed task-specific extrapolation coefficient. When $\alpha_k(s)=0$, the target reduces to the original teacher velocity and recovers standard OPD. When $\alpha_k(s)>0$, the target moves beyond the teacher along the base-to-teacher direction.

The output objective is then defined as
\begin{equation}
\begin{aligned}
\mathcal{L}_{\mathrm{out}}^{\rho}
&=
\mathbb{E}_{\substack{
k\sim\pi,\;
c\sim\mathcal{D}_k\\
\tau\sim p_{\theta}(\tau\mid c),\;
I\sim\rho_k(\cdot\mid c,\tau)
}}
\Biggl[
w_I
\Bigl\|
\mathbf{v}_{S_{\theta}}
\left(
\bar{\mathbf{z}}_I,t_I,c
\right)
\\[-0.5ex]
&\hspace{8.5em}
-
\operatorname{sg}
\left[
\widetilde{\mathbf{v}}_{k,I}(s)
\right]
\Bigr\|_2^2
\Biggr].
\end{aligned}
\label{eq:output_loss}
\end{equation}

The base model and teachers remain frozen, and gradients are propagated only through the student. Because different tasks may support different extrapolation strengths, we use a separately selected coefficient for each task.

To avoid using an aggressive target before the student has learned the teacher field, the effective coefficient is gradually activated during early training:
\begin{equation}
\alpha_k(s)
=
r(s)\alpha_k^{\max},
\qquad
r(s)\in[0,1],
\label{eq:alpha_warmup}
\end{equation}
where $s$ is the optimization step and $r(s)$ increases linearly from zero to one during a short warm-up period. After warm-up, $\alpha_k(s)$ remains fixed at $\alpha_k^{\max}$.

This design turns the teacher from a fixed learning endpoint into a task-specific direction for further improvement. It strengthens the changes introduced by task-specific post-training without directly scaling the entire teacher prediction. As a result, the student receives explicit supervision beyond teacher matching without changing the original on-policy rollout and procedure.

\input{table/main_table}

\subsection{Intermediate Representation Change Alignment}

Output supervision specifies the velocity that the student should predict, but it does not directly constrain how representations are transformed across the network. We therefore use the teacher's intermediate representation changes to guide the student's internal computation as shown in the left panel of Fig.~\ref{fig:method_overview}.

Let
$\mathbf{h}^{l}_{M, i} \in \mathbb{R}^{N \times D}$
denote the hidden representation produced by model $M$ after Transformer block $l$ at the $i$-th queried state, where $N$ is the number of tokens and $D$ is the hidden dimension. For an adjacent block pair
$p=(l_0,l_1)$, we define the inter-layer hidden-state change as
\begin{equation}
\Delta \mathbf{h}_{M, i}^{p}
=
\mathbf{h}_{M, i}^{l_1}
-
\mathbf{h}_{M, i}^{l_0},
\qquad
M \in \{S_{\theta},T_k\}.
\label{eq:hidden_delta}
\end{equation}

Rather than matching absolute hidden states, this quantity describes how a pair of blocks updates the current representation. We use non-overlapping adjacent block pairs that jointly cover all Transformer blocks.

We align both the direction and magnitude of these representation changes. Let
$\operatorname{Norm}(\cdot)$ denote token-wise RMS normalization over the hidden dimension. The direction loss:
\begin{equation}
\begin{aligned}
\ell_{\mathrm{dir}}^{k,i}
&=
\frac{1}{|\mathcal{P}|}
\sum_{p \in \mathcal{P}}
\operatorname{MSE}
\Bigl(
\operatorname{Norm}
\left(
\Delta \mathbf{h}_{S_{\theta},i}^{p}
\right),
\\[-0.4ex]
&\hspace{7em}
\operatorname{sg}
\left[
\operatorname{Norm}
\left(
\Delta \mathbf{h}_{T_k,i}^{p}
\right)
\right]
\Bigr).
\end{aligned}
\label{eq:hidden_direction}
\end{equation}
where $\mathcal{P}$ is the set of selected adjacent block pairs. This loss encourages the student and teacher to update each token representation in similar directions.

To align the update scale, we define
$m_{M}^{p}=\operatorname{RMS}(\Delta\mathbf{h}_{M}^{p})$,
where the RMS is computed over the token and hidden dimensions. The magnitude loss is
\begin{equation}
\begin{aligned}
\ell_{\mathrm{mag}}^{k,i}
&=
\frac{1}{|\mathcal{P}|}
\sum_{p \in \mathcal{P}}
\Biggl[
\log
\left(
m_{S_{\theta},i}^{p}+\epsilon
\right)
\\[-0.4ex]
&\hspace{7em}
-
\operatorname{sg}
\left[
\log
\left(
m_{T_k,i}^{p}+\epsilon
\right)
\right]
\Biggr]^2.
\end{aligned}
\label{eq:hidden_magnitude}
\end{equation}

where $\epsilon$ is a small constant for numerical stability. The hidden-state change objective is
\begin{equation}
\ell_{\mathrm{hidden}}^{k,i}
=
\ell_{\mathrm{dir}}^{k,i}
+
\beta\,
\ell_{\mathrm{mag}}^{k,i},
\label{eq:local_hidden_loss}
\end{equation}
where $\beta$ balances direction and magnitude alignment.

We evaluate this objective at the same on-policy states selected by the
underlying OPD query strategy. The resulting hidden-state objective is
\begin{equation}
\begin{aligned}
\mathcal{L}_{\mathrm{hidden}}^{\rho}
&=
\mathbb{E}_{\substack{
k\sim\pi,\;
c\sim\mathcal{D}_k\\
\tau\sim p_{\theta}(\tau\mid c)
}}
\left[
\sum_{i=0}^{N-1}
\rho_k(i\mid c,\tau)\,
\ell_{\mathrm{hidden}}^{k,i}
\right]
\\
&=
\mathbb{E}_{\substack{
k\sim\pi,\;
c\sim\mathcal{D}_k\\
\tau\sim p_{\theta}(\tau\mid c),\;
I\sim\rho_k(\cdot\mid c,\tau)
}}
\left[
\ell_{\mathrm{hidden}}^{k,I}
\right].
\end{aligned}
\label{eq:hidden_loss}
\end{equation}

Importantly, the hidden-state changes are matched directly to the teacher and are not extrapolated using the base model. Thus, the output target is allowed to move beyond the teacher, while the student's internal representation changes remain aligned with the teacher. The final training objective is
\begin{equation}
\mathcal{L}^{\rho}
=
\mathcal{L}_{\mathrm{out}}^{\rho}
+
\lambda_{\mathrm{h}}\,
\mathcal{L}_{\mathrm{hidden}}^{\rho},
\label{eq:total_objective}
\end{equation}
where $\lambda_{\mathrm{h}}$ controls the contribution of hidden-state change alignment. We use the same $\lambda_{\mathrm{h}}$ for all tasks.

%% file: table/main_table.tex
\begin{table*}[t]
    \centering

    \small
    \setlength{\tabcolsep}{3.5pt}
    \renewcommand{\arraystretch}{1.08}

    \resizebox{\textwidth}{!}{%
    \begin{tabular}{lcccccccccccc}
        \toprule

        \multirow{2}{*}{\textbf{Model}}
        & \multicolumn{7}{c}{\textbf{GenEval}}
        & \multirow{2}{*}{\textbf{OCR}}
        & \multirow{2}{*}{\textbf{PickScore}}
        & \multirow{2}{*}{\textbf{Aesthetic}}
        & \multirow{2}{*}{\textbf{HPSv2.1}}
        & \multirow{2}{*}{\textbf{ImgRwd}} \\

        \cmidrule(lr){2-8}

        & \textbf{Sing}
        & \textbf{Two}
        & \textbf{Cnt}
        & \textbf{Col}
        & \textbf{Pos}
        & \textbf{C-Attr}
        & \textbf{Overall}
        &
        &
        &
        &
        & \\

        \midrule

        \grouphead{a}{Base Models}
        FLUX.1-Dev
        & 0.778
        & 0.427
        & 0.628
        & 0.649
        & 0.068
        & 0.253
        & 0.467
        & 0.496
        & 22.648
        & 5.737
        & 0.292
        & 0.796 \\

        SD3.5-L
        & 0.703
        & 0.389
        & 0.525
        & 0.628
        & 0.140
        & 0.265
        & 0.442
        & 0.635
        & 21.973
        & 5.299
        & 0.264
        & 0.688 \\

        SD3.5-M+CFG
        & 0.869
        & 0.495
        & 0.569
        & 0.715
        & 0.148
        & 0.355
        & 0.525
        & 0.559
        & 22.398
        & 5.389
        & 0.280
        & 0.853 \\

        \midrule

        \grouphead{b}{Single-Task Teachers}

        SD3.5-M+GenEval Teacher
        & \second{0.994}
        & 0.952
        & 0.944
        & 0.923
        & 0.962
        & 0.833
        & 0.935
        & 0.401
        & 22.069
        & 5.247
        & 0.249
        & 0.614 \\

        SD3.5-M+OCR Teacher
        & 0.872
        & 0.472
        & 0.594
        & 0.723
        & 0.145
        & 0.355
        & 0.527
        & 0.938
        & 22.282
        & 5.256
        & 0.272
        & 0.918 \\

        SD3.5-M+Aes Teacher
        & 0.744
        & 0.422
        & 0.588
        & 0.676
        & 0.163
        & 0.323
        & 0.486
        & 0.576
        & \second{24.031}
        & 6.220
        & \second{0.346}
        & 1.504 \\

        \midrule

        \grouphead{c}{Diffusion/Flow RL and Multi-Objective OPD Methods}
        
        Flow-GRPO
        & \best{1.000}
        & \second{0.972}
        & 0.950
        & 0.924
        & 0.965
        & \second{0.845}
        & 0.943
        & 0.924
        & 23.500
        & 5.922
        & 0.321
        & 1.346
        \\

        DiffusionNFT
        & \best{1.000}
        & 0.968
        & \second{0.954}
        & 0.937
        & 0.963
        & 0.818
        & 0.940
        & 0.910
        & 23.800
        & 6.010
        & 0.331
        & 1.490
        \\

        DanceOPD
        & 0.975
        & 0.939
        & 0.934
        & 0.904
        & 0.940
        & 0.803
        & 0.916
        & 0.943
        & 23.836
        & 6.159
        & 0.328
        & 1.423 \\

        \textbf{+ \mymethod (Ours)}
        & \best{1.000}
        & 0.960
        & 0.950
        & \second{0.941}
        & \second{0.970}
        & \second{0.845}
        & \second{0.944}
        & \best{0.957}
        & 23.994
        & \second{6.297}
        & 0.344
        & \second{1.518} \\

        DiffusionOPD
        & \best{1.000}
        & 0.962
        & 0.938
        & 0.926
        & 0.948
        & 0.790
        & 0.927
        & 0.941
        & 23.941
        & 6.208
        & 0.340
        & 1.503 \\

        \textbf{+ \mymethod (Ours)}
        & \best{1.000}
        & \best{0.990}
        & \best{0.988}
        & \best{0.957}
        & \best{0.980}
        & \best{0.853}
        & \best{0.961}
        & \second{0.946}
        & \best{24.053}
        & \best{6.321}
        & \best{0.349}
        & \best{1.533} \\

        \bottomrule
    \end{tabular}%
    }

    \caption{\textbf{Quantitative comparison results on compositional alignment, text rendering, and human preference.} We compare STEP-OPD with representative base models, task-specific teachers, diffusion/flow RL methods and multi-objective OPD baselines at $512\times512$ resolution. STEP-OPD consistently improves both DanceOPD and DiffusionOPD across GenEval, OCR, and preference-based metrics, while also outperforming representative RL-based post-training methods. More importantly, it further surpasses the corresponding single-task teachers across all three capability groups. The best and second-best results are highlighted in \textbf{bold} and \underline{underline}, respectively.}
    
    \label{tab:main_comparison}
\end{table*}

%% file: section/experiments.tex
\section{Experiments}

\begin{figure*}[t]
    \centering
    \includegraphics[width=1.0\linewidth]{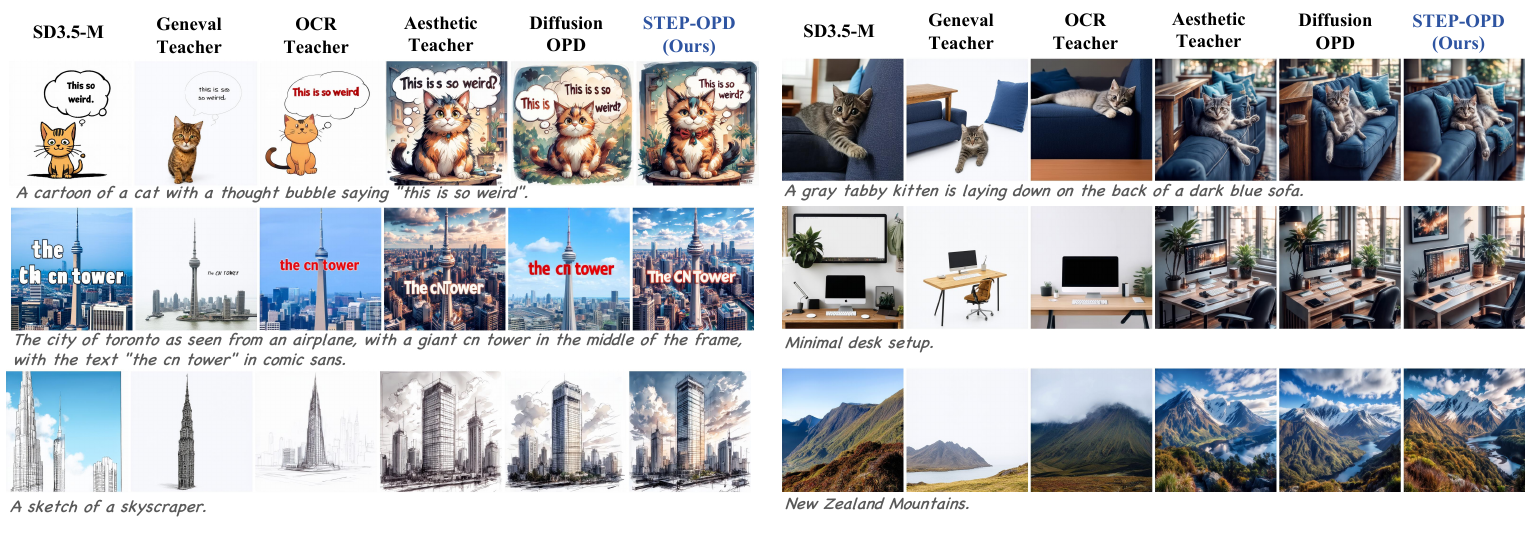}
   \caption{\textbf{Qualitative comparison} of SD3.5-M, task-specific teachers, DiffusionOPD, and STEP-OPD. The selected prompts cover the three evaluated capability groups: compositional alignment, text rendering, and human preference. Each row presents outputs generated from the same prompt. Across these diverse settings, STEP-OPD effectively consolidates the strengths of specialized teachers, achieving more accurate prompt adherence and text rendering while maintaining strong visual quality.}
    \label{fig:view}
\end{figure*}

\subsection{Experimental Setup}

\noindent\textbf{Models and Evaluation.}
Following the experimental setting of DiffusionOPD~\cite{li2026diffusionopd}, we use Stable Diffusion 3.5 Medium (SD3.5-M)~\cite{esser2024scaling} as the shared base model and student backbone, and train all models at a resolution of $512\times512$. Three task-specific teachers, all derived from the same base model, are used for compositional alignment, text rendering, and human preference, respectively. For preference-oriented training, we use prompts from Pick-a-Pic~\cite{kirstain2023pick}. We evaluate compositional alignment on GenEval~\cite{ghosh2023geneval}, reporting single object, two objects, counting, color, position, color attribution, and the overall score. Text rendering is evaluated using OCR accuracy. Human preference and visual quality are evaluated on DrawBench~\cite{saharia2022photorealistic} using PickScore~\cite{kirstain2023pick}, Aesthetic Score, HPSv2.1~\cite{wu2023human}, and ImageReward~\cite{xu2023imagereward}. All compared methods use the same evaluation prompts, random seeds, resolution, sampler, and guidance settings.

\medskip


\noindent\textbf{Baselines.}
We compare with three groups of baselines. First, we include representative pretrained image generators, including FLUX.1-Dev~\cite{flux2024}, SD3.5-Large, and SD3.5-Medium with classifier-free guidance~\cite{ho2022classifier}. Second, we report the three specialized teachers to measure whether the unified student can surpass their task-specific capabilities. Third, we consider representative post-training methods for diffusion and flow models include Flow-GRPO~\cite{liu2025flowgrpotrainingflowmatching} and DiffusionNFT~\cite{zheng2026diffusionnftonlinediffusionreinforcement}. We compare with DanceOPD~\cite{zhou2026danceopd} and DiffusionOPD~\cite{li2026diffusionopd}, two representative OPD frameworks for image generation. We apply our method to both OPD baselines using the same student initialization, teacher models, and evaluation settings. 

\medskip

\noindent\textbf{Implementation Details.}
All experiments are conducted at a resolution of $512\times512$. The student generates $10$-step on-policy trajectories with a deterministic flow solver, and only LoRA~\cite{hu2022lora} parameters are optimized. For a fair comparison, all methods are trained for $1000$ epochs on 8 NVIDIA A100 GPUs. We use task-specific extrapolation coefficients of $0.01$, $0.06$, and $0.20$ for GenEval, OCR, and aesthetics, with a short linear warm-up. Hidden-state change alignment is applied to $12$ adjacent block pairs covering all $24$ Transformer blocks, with $\lambda_{\mathrm{h}}=0.005$ and $\beta=0.1$.

\subsection{Main Results}

\input{table/ablation_method}

\noindent\textbf{Quantitative Results.}
Table~\ref{tab:main_comparison} presents the quantitative results on compositional alignment, text rendering, and human preference. STEP-OPD consistently improves both DanceOPD and DiffusionOPD across all three capability groups. When applied to DiffusionOPD, it increases the GenEval overall score from $0.927$ to $0.961$ and the OCR score from $0.941$ to $0.946$, while also improving all four preference-based metrics. Both STEP-OPD variants further outperform the representative diffusion/flow RL methods, demonstrating consistent advantages over both OPD and RL based post-training methods.

More importantly, the DiffusionOPD-based variant surpasses all corresponding single-task teachers, achieving $0.961$ versus $0.935$ on GenEval, $0.946$ versus $0.938$ on OCR, and higher scores on all four preference metrics. This demonstrates that STEP-OPD generalizes across OPD frameworks and enables beyond-teacher learning.

\medskip

\noindent\textbf{Qualitative Comparison.}
Figure~\ref{fig:view} compares SD3.5-M, three task-specific teachers, DiffusionOPD, and STEP-OPD on prompts covering compositional alignment, text rendering, and human preference. For compositional prompts, STEP-OPD more faithfully captures the requested objects, attributes, and spatial relations, as illustrated by the gray tabby kitten lying on the back of a dark blue sofa. For text-intensive prompts, it renders the specified phrases more accurately, as shown in the thought-bubble and CN Tower examples. It also produces more detailed, coherent, and visually appealing results for preference-oriented prompts. Together with the quantitative and ablation results, they demonstrate effective capability consolidation and beyond-teacher learning.

\subsection{Analysis}


\noindent\textbf{Component Ablation.}
Table~\ref{tab:component_ablation} examines the contributions of output extrapolation and intermediate representation change alignment. Representation change alignment alone consistently improves compositional alignment and text rendering, increasing GenEval from $0.927$ to $0.959$ and OCR from $0.941$ to $0.944$. In contrast, output extrapolation brings larger gains on the preference-based metrics while also improving GenEval and OCR, showing that a direct evidence that the extrapolated output target enables
systematic beyond-teacher learning. Combining the two components enables the student to learn the preference-oriented capabilities while further improving compositional alignment, text rendering. 

\begin{figure}
    \centering
    \includegraphics[width=1.0\linewidth]{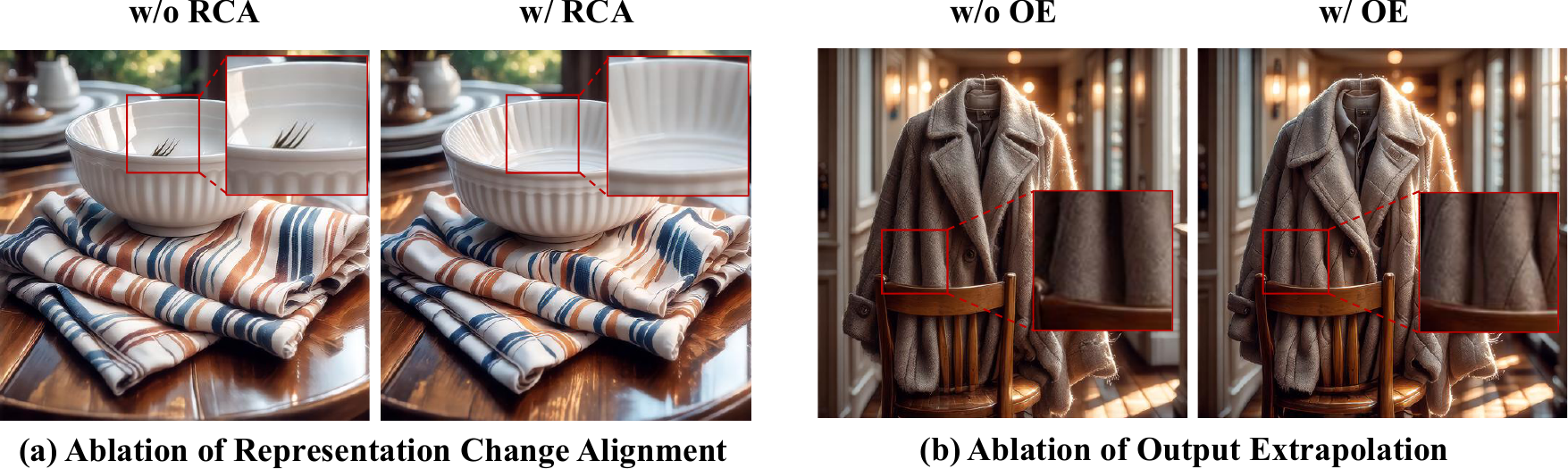}
        \caption{
        \textbf{Qualitative ablation of representation change alignment
        (RCA) and output extrapolation (OE).}
        Each pair is generated using the same prompt and random seed.
    }
    \label{fig:ablation_view}
\end{figure}

Figure~\ref{fig:ablation_view} provides complementary qualitative evidence: without RCA, the bowl exhibits spurious internal structures and inconsistent local geometry, whereas RCA produces a cleaner and more structurally coherent shape. Without OE, the coat shows weaker fabric texture and a flatter material appearance, while OE yields richer fine-grained details and stronger overall visual quality.

\medskip

\begin{figure}
    \centering
    \includegraphics[width=1.0\linewidth]{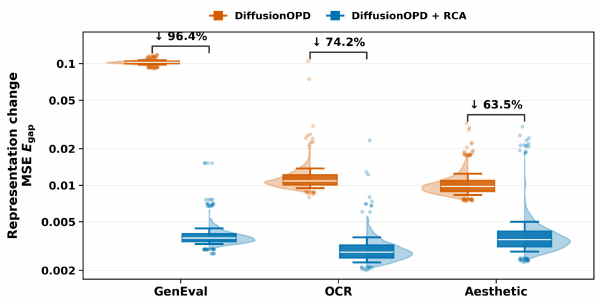}
    \caption{ \textbf{Representation change gap distributions with and without RCA.} Compared with DiffusionOPD, adding RCA consistently shifts the discrepancy distributions toward lower values across GenEval, OCR, and Aesthetic. Annotations give the median paired relative reduction of ${E}_{\mathrm{gap}}$ under RCA. Lower values indicate closer alignment with the teacher's internal representation changes.}
    \label{fig:rca_gap}
\end{figure}

\noindent\textbf{Effect of Extrapolation Strength.}
Table~\ref{tab:alpha_ablation} shows that the preferred extrapolation strength varies across tasks. GenEval performs best at $\alpha_G=0.01$, suggesting that compositional alignment is particularly sensitive to target displacement along the base-to-teacher direction, since aggressive extrapolation may disrupt the balance among object identity, count, color, spatial relations, and attribute binding. OCR favors an intermediate coefficient of $\alpha_O=0.06$, whereas the preference task benefits from a larger value of $\alpha_A=0.20$, indicating that its broader perceptual improvement direction remains informative over a longer range. Performance generally declines beyond the task-specific optimum, showing that extrapolation is effective within a reliable neighborhood of the teacher direction rather than improving monotonically with strength. These results support using task-specific coefficients instead of a single shared value.

\input{table/alpha_ablation}

\medskip

\noindent\textbf{Representation Change Gap Analysis.}
To verify that RCA improves the student's internal transformations rather than merely acting as a regularizer, we measure the discrepancy between the student and teacher representation changes. For each matched prompt and seed pair, we compute
$\mathcal{E}_{\mathrm{gap}}
=
\mathcal{E}_{\mathrm{dir}}
+
0.1\mathcal{E}_{\mathrm{mag}}$
over the selected adjacent block pairs and average it across the
$10$-step on-policy trajectory. As shown in
Figure~\ref{fig:rca_gap}, RCA shifts the gap distributions toward lower
values, with median paired relative reductions of $96.4\%$, $74.2\%$,
and $63.5\%$ on GenEval, OCR, and Aesthetic, respectively. This confirms
that output-level velocity matching leaves blockwise representation
evolution underconstrained, whereas RCA enables the student to learn the
teacher's inter-block transformation dynamics. Together with the gains
in Fig.~\ref{fig:teaser} and Table~\ref{tab:component_ablation}, these
reduced gaps support more effective transfer of compositional and
text-rendering capabilities.

%% file: table/ablation_method.tex










\begin{table}[t]
    \centering
    \small
    \setlength{\tabcolsep}{3.2pt}
    \renewcommand{\arraystretch}{1.08}

    \resizebox{\columnwidth}{!}{%
    \begin{tabular}{@{}lcccccc@{}}
        \toprule
        \textbf{Model}
        & \textbf{GenEval}
        & \textbf{OCR}
        & \textbf{PickScore}
        & \textbf{Aesthetic}
        & \textbf{HPSv2.1}
        & \textbf{ImgRwd} \\
        \midrule

        DiffusionOPD
        & 0.927
        & 0.941
        & 23.941
        & 6.208
        & 0.340
        & 1.503 \\

        + RCA
        & \second{0.959}
        & \second{0.944}
        & 23.980
        & 6.217
        & 0.343
        & 1.516 \\

        + OE
        & 0.957
        & 0.943
        & \second{24.046}
        & \second{6.312}
        & \second{0.348}
        & \second{1.530} \\

        \textbf{+ \mymethod (Ours)}
        & \best{0.961}
        & \best{0.946}
        & \best{24.053}
        & \best{6.321}
        & \best{0.349}
        & \best{1.533} \\

        \bottomrule
    \end{tabular}%
    }

    \caption{\textbf{Component ablation of STEP-OPD.} \textbf{OE} denotes output extrapolation, and
        \textbf{RCA} denotes representation change alignment. The best and second-best results are shown in \textbf{bold} and \underline{underlined}, respectively.}
    \label{tab:component_ablation}
\end{table}

%% file: table/alpha_ablation.tex
\begin{table}[t]
    \centering

    \scriptsize
    \setlength{\tabcolsep}{2.8pt}
    \renewcommand{\arraystretch}{1.00}

    \resizebox{\columnwidth}{!}{%
    \begin{tabular}{lcccccc}
        \toprule

        \textbf{Coefficient}
        & \textbf{GenEval}
        & \textbf{OCR}
        & \textbf{PickScore}
        & \textbf{Aesthetic}
        & \textbf{HPSv2.1}
        & \textbf{ImgRwd} \\

        \midrule

        \multicolumn{7}{l}{
            \textbf{(a) GenEval extrapolation coefficient $\alpha_G$}
        } \\

        \selectedcoef{0.01}
        & \best{0.961}
        & \best{0.946}
        & \best{24.053}
        & \best{6.321}
        & \best{0.349}
        & \best{1.533} \\

        0.03
        & \second{0.956}
        & \second{0.941}
        & 24.040
        & \second{6.302}
        & \second{0.347}
        & \second{1.527} \\

        0.05
        & 0.955
        & \second{0.941}
        & \second{24.052}
        & 6.299
        & \second{0.347}
        & 1.525 \\

        0.07
        & 0.951
        & 0.938
        & 24.038
        & 6.295
        & 0.346
        & 1.517 \\

        \midrule

        \multicolumn{7}{l}{
            \textbf{(b) OCR extrapolation coefficient $\alpha_O$}
        } \\

        0.02
        & 0.956
        & 0.939
        & 24.040
        & 6.300
        & \second{0.347}
        & 1.526 \\

        0.04
        & \second{0.959}
        & 0.942
        & \second{24.041}
        & \second{6.316}
        & \second{0.347}
        & \second{1.532} \\

        \selectedcoef{0.06}
        & \best{0.961}
        & \best{0.946}
        & \best{24.053}
        & \best{6.321}
        & \best{0.349}
        & \best{1.533} \\

        0.08
        & \second{0.959}
        & \second{0.945}
        & 24.036
        & 6.297
        & \second{0.347}
        & 1.531 \\

        \midrule

        \multicolumn{7}{l}{
            \textbf{(c) Aesthetic extrapolation coefficient $\alpha_A$}
        } \\

        0.10
        & 0.957
        & 0.940
        & 24.024
        & 6.265
        & 0.345
        & 1.514 \\

        0.15
        & 0.959
        & 0.942
        & \second{24.047}
        & 6.281
        & 0.346
        & 1.528 \\

        \selectedcoef{0.20}
        & \best{0.961}
        & \best{0.946}
        & \best{24.053}
        & \best{6.321}
        & \best{0.349}
        & \best{1.533} \\

        0.25
        & \second{0.960}
        & \second{0.943}
        & 24.033
        & \second{6.320}
        & \second{0.348}
        & \second{1.532} \\

        \bottomrule
    \end{tabular}%
    }
    \caption{
        \textbf{Sensitivity analysis of the extrapolation coefficients.} Best results are shown in bold and second-best results
        are underlined within each subsection. \(\dagger\) denotes the task-specific coefficient adopted in all main experiments.
    }
    \label{tab:alpha_ablation}
\end{table}

%% file: section/conclusion.tex
\section{Conclusion}

We presented \textbf{\mymethod}, an OPD framework for image generation that addresses the limitations of an objective restricted to teacher matching and output-only supervision. Our method constructs targets beyond the teacher by extrapolating the velocity change from the shared base model to each task-specific teacher, while aligning the direction and magnitude of inter-layer representation changes to guide the student's internal transformations. Experiments on compositional alignment, text rendering, and human preference show consistent improvements over standard OPD methods. The resulting unified student surpasses the corresponding single-task teachers across all three capability groups. 